\pdfoutput=1

\documentclass[11pt]{article}

\usepackage[dvipsnames]{xcolor}

\usepackage[final]{acl}

\usepackage{times}
\usepackage{latexsym}

\usepackage[T1]{fontenc}

\usepackage[utf8]{inputenc}

\usepackage{microtype}

\usepackage{inconsolata}

\usepackage{graphicx}

\usepackage{subfigure}
\usepackage{booktabs} 
\usepackage{xspace}
\usepackage{wrapfig}
\usepackage[para,online,flushleft]{threeparttable}
\usepackage{hyperref}

\usepackage{amsmath}
\usepackage{amssymb}
\usepackage{mathtools}
\usepackage{amsthm}
\usepackage{listings}

\usepackage[textsize=tiny]{todonotes}

\title{Beyond instruction-conditioning, \\MoTE: Mixture of Task Experts for Multi-task Embedding Models}

\author{
 \textbf{Miguel Romero\textsuperscript{1,2, \footnotemark[1]}},
 \textbf{Shuoyang Ding\textsuperscript{1}},
\\
 \textbf{Corey D. Barret\textsuperscript{1}},
 \textbf{Georgiana Dinu\textsuperscript{1}},
 \textbf{George Karypis\textsuperscript{1,2, \footnotemark[1]}}
\\
\\
 \textsuperscript{1}Amazon,
 \textsuperscript{2}University of Minnesota
}

\begin{document}

\def\mote{MoTE\xspace}
\def\tacl{TA-CL\xspace}
\def\ea{EA\xspace}

\maketitle
\footnotetext[1]{Corresponding authors, emails: romer333@umn.edu, karypis@umn.edu}
\begin{abstract}
Dense embeddings are fundamental to modern machine learning systems, powering Retrieval-Augmented Generation (RAG), information retrieval, and representation learning. While instruction-conditioning has become the dominant approach for embedding specialization, its direct application to low-capacity models imposes fundamental representational constraints that limit the performance gains derived from specialization. In this paper, we analyze these limitations and introduce the Mixture of Task Experts (\mote) transformer block, which leverages task-specialized parameters trained with Task-Aware Contrastive Learning (\tacl) to enhance the model ability to generate specialized embeddings. Empirical results show that \mote achieves $64\%$ higher performance gains in retrieval datasets ($+3.27 \rightarrow +5.21$) and $43\%$ higher performance gains across all datasets ($+1.81 \rightarrow +2.60$). Critically, these gains are achieved without altering instructions, training data, inference time, or number of active parameters.
\end{abstract}

\section{Introduction}


\begin{figure*}[ht]
    \centering
    \includegraphics[width=0.89\textwidth]{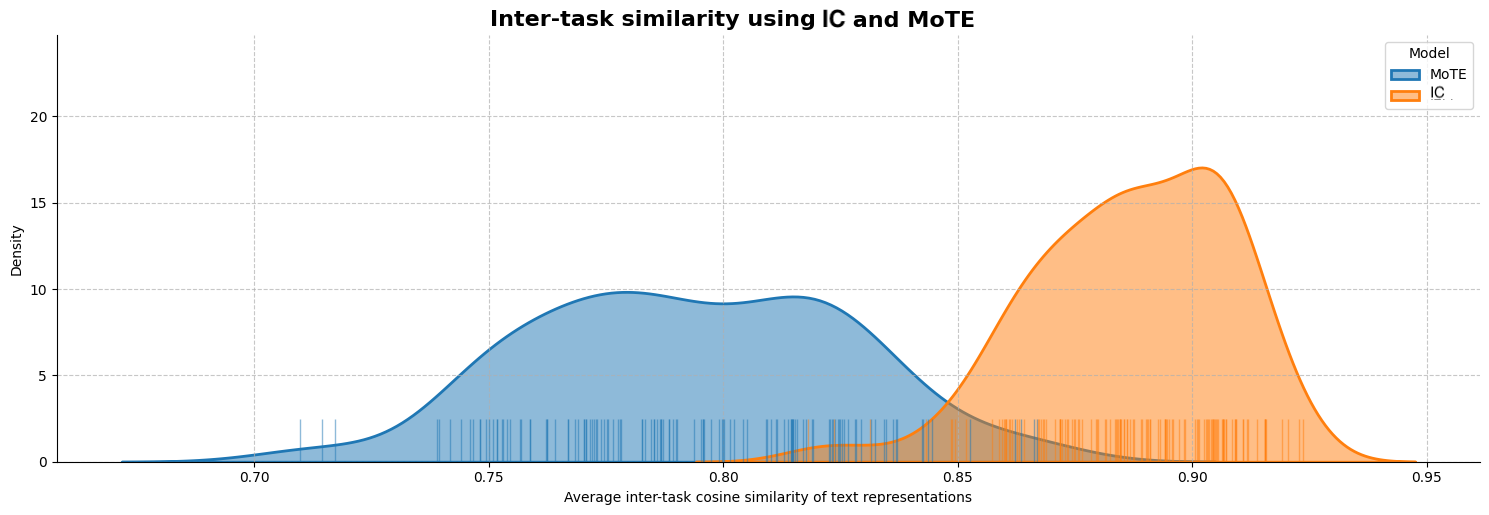} 
    \caption{Average inter-task cosine similarity between embedding representations for the same sequence. Results show that compared to Instruction-Conditioning (IC), Mixture-of-Task Experts (\mote) provides more flexibility (lower cosine-similarity) generating specialized embeddings. 
    }
    \label{fig:task-diverging-distribution}
\end{figure*}

Semantic text representations are a key component to many real-world applications such as search, recommendation systems, and spam classification. These representations are commonly obtained using embedding models that map unstructured text into a dense $n$-dimensional vector referred to as dense embedding. Contemporary embedding models are trained by simultaneously optimizing the model towards a wide range of downstream tasks including Retrieval Augmented Generation (RAG) \citep{gao2023retrieval, lewis2020retrieval}, search \citep{wise2020covid}, Semantic Text Similarity (STS) \citep{chandrasekaran2021evolution}, classification \citep{oneill-etal-2021-wish, wang2021lm}, and clustering \citep{xu2015short}. As a result, a single embedding model is able to generate a dense text representation that can be used in all these downstream tasks.

Historically, dense embedding models have been trained using a multi-task curriculum, where a single model is optimized across multiple datasets, each corresponding to a specific task such as clustering or retrieval. In this framework, a shared embedding space is used across all tasks, allowing a single representation to serve multiple downstream applications. However, different tasks impose distinct semantic similarity requirements, often leading to conflicting objectives. For instance, in a semantic textual similarity (STS) task, the questions ``Who was Isaac Newton?" and ``Who was the father of Calculus?" should be close in the embedding space since they refer to the same person. In contrast, a retrieval task requires these queries to be far apart, as they can not answer each other \citep{gao2021simcse}. This presents a fundamental challenge in embedding model development, where a single embedding space must be carefully tuned to support multiple tasks without compromising performance on any of them \citep{muennighoff2022mteb, neelakantan2022text}. To address this, modern embedding models have shifted\footnote{\url{https://huggingface.co/spaces/mteb/leaderboard}} toward generating task-specific embeddings, allowing representations to be adapted based on the downstream application. The consistent performance gains observed from embedding specialization, coupled with its widespread adoption in state-of-the-art models \citep{wang2022text,su2022one} have provide strong evidence of the limitations of generic embeddings in downstream tasks.

The predominant approach for embedding specialization is Instruction-Conditioning (IC) \cite{wang2022text,su2022one}, where the input text is enriched with task-specific instructions to steer the generation of specialized embeddings. For instance, \citet{wang2022text} employs the instruction "query:" for queries and "passage:" for passages in retrieval tasks, providing the model with information to distinguish between different roles in the retrieval tasks. With this approach, the specialization signal is entirely encoded in the instruction tokens, which propagate through the model to produce the final embeddings.

However, theoretical properties of neural networks, such as Lipschitz continuity, suggest that small changes in input—like adding short instruction tokens—produce only minor shifts in output embeddings \citep{tang2024understanding}, limiting a model’s ability to create disentangled, task-specific representations. This constraint is amplified in smaller models and with short prompts, where low-dimensional embeddings face capacity bottlenecks \citep{tishby2000information}, forcing trade-offs between preserving semantic and task-relevant information. These limitations are compounded during multi-task contrastive training, where gradient interference \citep{yu2020gradient} and conflicting objectives \citep{ravi2020navigating} saturate model capacity, and where hyper-parameters like batching strategy and contrastive temperature must be carefully tuned to align with downstream task semantics, further complicating effective task-specific embedding specialization.



In this paper, we study these limitations in the context of multi-task embedding models and propose alternative methods to enhance embedding specialization. Our key contributions include:

\begin{enumerate}
    \item We provide empirical evidence that relying solely on instruction-conditioning can constrain the adaptability of embeddings across diverse tasks. We analyze task-specific performance trade-offs and demonstrate how instruction-conditioned models struggle to fully disentangle task representations in low-capacity settings (Figure \ref{fig:task-diverging-distribution}).
    \item We introduce \textit{Mixture of Task Experts (\mote)}, a Mixture of Experts (MoE) block with task-specialized experts.
    \item We propose \textit{Task-Aware Contrastive Learnign (\tacl)}, a novel training curriculum to tailor the training of \mote's experts to their associated downstream task.
\end{enumerate}

Experiments conducted on 56 datasets across 7 tasks \citep{muennighoff2022mteb} demonstrate that leveraging \mote and \tacl enables more effective utilization of task information for embedding specialization (Figure \ref{fig:task-diverging-distribution}). This approach leads to $50\%$ higher performance gains in critical tasks such as retrieval and $31\%$ higher gains in other tasks compared to instruction-conditioned specialization. Critically, these improvements are achieved with identical input information, training data, latency, and number of active parameters. Furthermore, exploiting the observation that embedding models are typically used for one downstream task at a time (e.g., retrieval-indexing, retrieval-querying, classification), \mote can maintain the same GPU memory footprint as instruction-conditioned models by offloading inactive task experts to CPU or disk. This ensures computational efficiency without sacrificing performance.




\section{Background}

\subsection{Text Embedding Models}\label{subsect: foundational text embedding models}

Embedding models $f_\Theta$ map elements from the sequence space $\mathcal{S}$ to $\mathbb{R}^N$. Typically initialized from encoder-only pre-trained checkpoints like BERT \citep{devlin2018bert} or RoBERTa \citep{liu2019roberta}, they are trained via contrastive learning on multiple datasets containing pairs of sequences $(a,p)$ that should be close to each other in the embedding space (e.g, query-passage pairs in retrieval, passage-passage pairs in clustering, etc.). In this context a common contrastive loss is InfoNCE \citep{oord2018representation}:
\begin{equation*}
\mathcal{L}(f_\Theta; a, p, P^*) = -\log{\frac{e^{\gamma(u_a, u_p)}}{\sum_{d\in\{p, P^*\}}e^{\gamma(u_a, u_d)}}}
\end{equation*}
\noindent where $P^*$ denotes the other positive in the mini-batch, $u_{s}$ is the embedding obtained by embedding the sequence $s$, and $\gamma$ is a similarity measure such as cosine-similarity \citep{HAN201239}. This optimizes $\Theta$ by pulling similar sequences closer and pushing dissimilar ones apart, with the definition of similar being task-dependent.

\subsection{Instruction-Conditioned Embeddings}\label{subsect: instruction embeddings}

IC is a technique \cite{wang2024multilingual,muennighoff2024generative, wang2023improving, zhang2023instruction} to generate specialized embeddings with a single model by concatenating to the input sequence a text string that describes the downstream use case. This concatenation happens during training and inference. Thus, IC models can be defined as functions $f^I_\Theta=f_\Theta \circ\phi$ over $\mathcal{I}\times\mathcal{S}$ of the set of instructions $\mathcal{I}$ and sequences $\mathcal{S}$ where $\phi$ is the concatenation operation in the sequence space and $f_\Theta$ is an embedding model. 

Methods leveraging this strategy vary in the level of detail provided in the instructions. Small models \citep{wang2022text} use a pre-defined set of instructions with task level information such such as ``query:" and ``passage:" in retrieval use cases. Large models \cite{su2022one} use a free-form set of instructions with dataset-specific information such as ``Represent the Amazon comment for classifying the sentence as positive or negative:".

\subsection{Mixture-of-Experts}\label{subsect: mixture of experts}

MoE has been a long-standing technique \citep{yuksel2012twenty} that proposes a divide-and-conquer approach by leveraging multiple expert modules and an routing mechanism to identify the most appropriate expert for any given input. A recent application of this approach is the transformer MoE block \citep{lepikhin2020gshard,jiang2024mixtral,gao2022parameter} which leverages multiple Multi-Layer Perceptrons (MLP) as experts and a learned token-level mechanism to route individual tokens through multiple experts. Specifically, for each token, the routing mechanism $g$ uses the information in the intermediate token representation to identify the subset of experts that should process the representation \citep{lepikhin2020gshard}. Replicas of the intermediate token representation are then dispatched to the selected experts where they are processed independently. Lastly, a pooling layer aggregates the experts' output to generate a unique representation for each of the tokens. During training, an auxiliary expert-balancing objective is leveraged to learn the routing mechanism while ensure a consistent training of and a balanced workload across the experts modules \citep{jiang2024mixtral,gao2022parameter}.




\section{Method}\label{sec: method}

While instruction-conditioning enables the generation of specialized embeddings $f^\mathcal{I}_\Theta(i, s)$, relying solely on the input instruction to steer the specialization of the embedding while the other parameters are shared can limit the models ability to generate specialized embeddings \citep{ravi2020navigating,yu2020gradient}. To overcome this constraints we propose to introduce task specialized parameters to increase the model's capacity to specialize embeddings while maintaining the overall number of active parameters. Formally, we propose a function $\Psi^I_{[\Theta_0|\Theta_{I}]}$ such that:

\begin{alignat*}{2}
  \Psi^I_{[\Theta_0|\Theta_{I}]}:I\times S&\xlongrightarrow{} \mathbb{R}^N \\
  (\mathbf{i}, \mathbf{s})&\longmapsto f^i_{[\Theta_0|\Theta_{i}]}(i, s) = v_i
\end{alignat*}

\noindent where 

\begin{alignat*}{2}
  f^i_{[\Theta_0|\Theta_{i}]}:\{i\}\times S&\xlongrightarrow{\phi} S \xlongrightarrow{f_{[\Theta_0|\Theta_{i}]}} \mathbb{R}^N \\
  (\mathbf{i}, \mathbf{s})&\longmapsto \mathbf{i}|\mathbf{s} \longmapsto v_i
\end{alignat*}

Consistent with IC, $\Psi^I_{[\Theta_0|\Theta_{I}]}$ accepts tuples $(i,s)$ where $i\in\mathcal{I}$ represents the task instruction providing contextual information about the downstream task (e.g., ``query: '', ``passage: '') and  $s$ denotes the input sequence. However, unlike standard instruction-conditioning, $\Psi^I_{[\Theta_0|\Theta_{I}]}$ dynamically selects the appropriate set of specialized parameters $\Theta_i$ to route the input sequence $[i|s]$ and enhancing task specialization.

\subsection{Architecture}\label{subsect: arch}

\begin{figure*}[tb]
    \centering
    \includegraphics[width=0.9\textwidth]{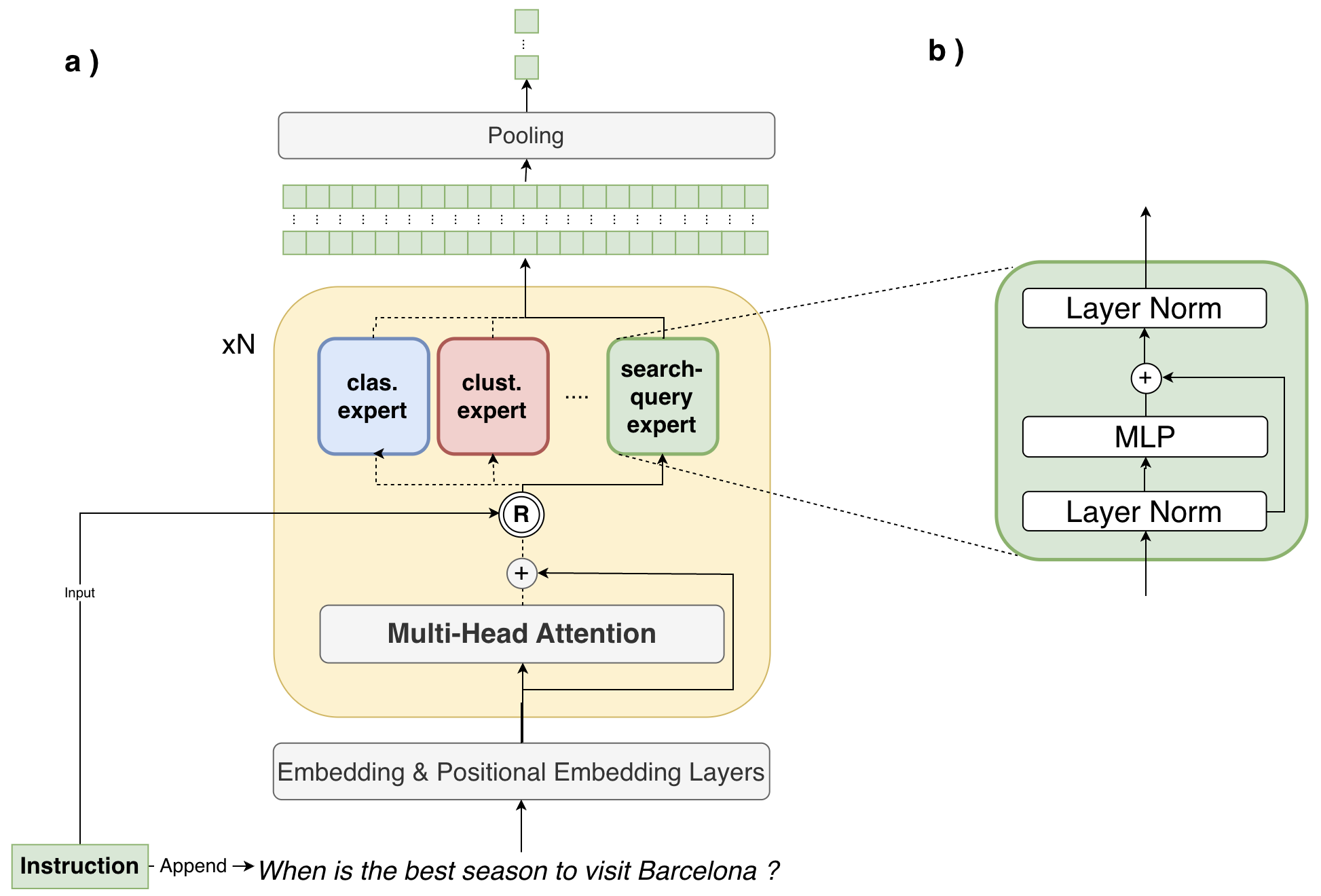}
    \caption{a) Overview of the \mote block, which replaces the standard transformer block in \mote. Each \mote block contains multiple task-specific experts and a routing mechanism that selects the appropriate expert based on the task instruction. b) Detailed view of a single expert’s internal structure, consisting of a layer normalization followed by a MLP network and a layer normalization, mirroring the standard transformer feedforward configuration but applied independently within each expert.}
    \label{fig:mote}
\end{figure*}

We implement $\Psi^I_{[\Theta_0|\Theta_{I}]}$ using the Mixture of Task Experts (\mote) transformer block (Figure \ref{fig:mote}a). \mote implements the specialized parameter sets $\Theta_{I}$ with different expert modules and leverages a task-based routing mechanism to process input sequences through  different experts based on its intended downstream use.

\mote's instruction-based routing mechanism $R: \mathcal{I} \rightarrow \mathcal{E}$ maps an instruction $i\in\mathcal{I}$ to an expert module $e\in\mathcal{E}$, which is then applied to the intermediate state of $[i|s]$. This approach offers two key benefits for embedding specialization. First, unlike traditional MoE routing, experts are trained exclusively on task-relevant examples, leading to more efficient training and better specialization. Second, \mote enhances the model’s capacity for embedding specialization without incurring additional latency costs by routing full sequences through a single expert rather than distributing tokens across multiple experts. Furthermore, in most practical applications, \mote avoids increased GPU memory usage, as inactive experts can be offloaded to the CPU or disk, ensuring efficient resource utilization.

Each expert block $e\in\mathcal{E}$ consists of a MultiLayer Perceptron (MLP) and two normalization layers (Figure \ref{fig:mote}b). The MLP enables specialized transformations for different tasks, while the normalization layers allow \mote to learn distinct centroids and variances for each task \cite{ba2016layer}, improving task adaptability.

\subsubsection{Initialization}\label{subsect: dense2moe}

\mote's task-routing mechanism requires task-augmented inputs to train the task experts. However, datasets associated with downstream embedding task are only available during contrastive learning while generic pre-training relies on generic task-agnostic sequences. 

To enable \mote to enhance the model’s specialization capacity while leveraging existing pre-trained checkpoints we up-cycle \citep{komatsuzaki2022sparse, he2024upcycling} the dense pre-trained blocks into \mote blocks before the contrastive learning stage. Specifically, we follow a three-step process. First, we select the dense transformer blocks to be converted into \mote blocks. Second, for each selected block, we instantiate $|\mathcal{E}|$ experts using the same MLP configuration as the original dense transformer block. Last, we initialize all experts in $\mathcal{E}$ with a copy of the dense transformer's MLP weights. This approach ensures that the model retains knowledge from its pre-trained checkpoint while gaining the flexibility of task-specialized experts.

\subsection{Training}\label{subsect: training}

With \mote's instruction-based routing, training examples are directed to their corresponding task experts, ensuring that task-specific gradients are only backpropagated through the relevant expert. However, different downstream tasks have distinct representation requirements, influencing training nuances such as batching strategy and contrastive temperature (Appendix \ref{sec:analysis}). This poses a challenge because contrastive learning relies on negatives from the same mini-batch to compute the loss, yet different experts require slightly different training configurations.

To overcome this challenge, we introduce Task-Aware Contrastive Learning (\tacl), which dynamically adjusts training configurations based on the selected expert. \tacl leverages a hierarchical data structure that organizes training samples by task and dataset. During mini-batch construction, a single task is selected, and examples are drawn from one or more of the task's datasets according to that task’s batching strategy. Additionally, task metadata is passed with the mini-batch to the training loop to adjust the tailor the contrastive objective to the task. When combined with \mote, this approach ensures that each task expert is trained with a configuration suited to its specific downstream task.

\section{Experimental Methodology}\label{sec:experimental setup}




We evaluate performance using the Massive Text Embedding Benchmark (MTEB) \citep{muennighoff2022mteb}, which spans 7 tasks and 56 datasets (see Table \ref{tab:MTEB distribution}). Following MTEB's evaluation protocol, we use the same metrics and procedures, reporting the average performance per task and overall across all tasks.

\begin{table}[tb]
  \centering
  \caption{Metrics and distribution of datasets across tasks in the MTEB benchmark.}
  \begin{threeparttable}
  \begin{tabular}{lcc}
    \hline
    \textbf{Task}  & \textbf{\# Datasets} & \textbf{Metric} \\
    \hline
    \hline
    Retrieval   &   15  &  NDCG@10\\
    Class. &  12   & Accuracy \\
    Clustering &  11   & V-Measure \\
    Re-ranking &  4   & MAP \\
    Pair Class. &  3  & AP\\
    STS &  10   & Spearman corr. \\
    Summ. & 1   & Spearman corr. \\
    \hline
    \hline
  \end{tabular}
  \begin{tablenotes}
      \small For more information about the metrics please refer to Appendix \ref{appendix: metrics}.
  \end{tablenotes}
  \end{threeparttable}
  \label{tab:MTEB distribution}
\end{table}

We evaluate the embedding specialization benefits derived from Instruction-Conditioning (IC) and \mote on \citet{nussbaum2024nomic}'s model\footnote{Within top-10 of the 100-250M parameter models as of 02/08/2025}. Both candidates are initialized using the same pre-trained checkpoint\footnote{\url{https://huggingface.co/nomic-ai/nomic-bert-2048}}, and trained on the same classification, clustering and retrieval datasets \footnote{\url{https://github.com/nomic-ai/contrastors/blob/main/src/contrastors/configs/data/contrastive_pretrain.yaml}} during contrastive learning \footnote{Dataset source: \url{https://huggingface.co/datasets/sentence-transformers/embedding-training-data}}. All candidates are trained for a single epoch using the InfoNCE objective \citep{oord2018representation} and the AdamW optimizer with a learning rate of $5\times10^{-6}$, a weight decay of $0.1$ and a batch size of $6,144$. 

The IC and \mote candidates leverage the same instructions: ``classification: '', ``clustering: '', ``search query: '', and ``search document: ''. Additionally, the \mote candidate is implemented by using \mote blocks in every transformer block and four experts in each transformer block, each of which is associated with one of the above instructions. Further implementation details are provided in Appendix \ref{app: further impl details}.

Additionally, to assess the performance benefits of multi-task embedding models with task-specific experts compared to specialized single-task models, we train and evaluate three Specialized Embedding Models (SEM). Each SEM uses the same underlying transformer architecture as the EM and IEM candidates but is trained independently on one of the classification, clustering, or retrieval subsets of the training data. Performance is then reported for each SEM on its corresponding downstream task. 

\section{Results}

\subsection{Performance across seen tasks}\label{subsection: exp arch perf in scope}

This section explores the performance of \mote in tasks that were used during training, 
namely retrieval, classification and clustering (we will refer to those as \emph{seen} tasks). Table \ref{tab:value of foundational model} shows the absolute average dataset performance of the generic embedding model as well as the relative gains derived from each of the specialization methods at the task level and across all seen tasks.

\begin{table}[tb]
  \centering
  \begin{threeparttable}
  \caption{Comparison of performance gains from different embedding specialization approaches.}
  \label{tab:value of foundational model}
  \begin{tabular}{lcccc}
    \hline
    \textbf{Task} & \textbf{EM} & \textbf{SEM} & \textbf{IC} & \textbf{\mote} \\
    \hline
    \hline
    Retrieval \tnote{1} & 42.31 & -8.86 & +3.27 & \textbf{+5.21} \\
    Class. \tnote{2} & 65.37 & -1.29 & +3.37 & \textbf{+3.79} \\
    Clustering \tnote{3} & 42.95 & -3.84 & \textbf{+0.14} & -0.06 \\
    \hline
    Average\tnote{4} & 49.78 & -5.02 & +2.35 & \textbf{+3.23}\\
    \hline
    \hline
  \end{tabular}
  \begin{tablenotes}
\small Comparison of performance gains of Specialized Embedding Models (SEM), Instruction-Conditioning (IC) and \mote with respect to the non-specialized Embedding Model (EM) on tasks seen during training. \item[1] NDCG@10; \item[2] accuracy; \item[3] validity measure; \item[4] Score average across 38 datasets. 
  \end{tablenotes}
  \end{threeparttable}
\end{table}

In retrieval, \mote achieves a +5.21 improvement, surpassing IC’s +3.27. Similarly, in classification, \mote shows a +3.79 gain compared to IC’s +3.37. Overall, \mote achieves the highest average dataset performance improvement (+3.23), exceeding IC’s +2.35, while maintaining latency and memory usage. Additionally, while \mote improves performance on clustering tasks relative to SEM, its gains are modest compared to other tasks. This appears to stem from the high similarity between search-document and clustering embedding spaces and is further investigated in Section \ref{subsect: dynamic training regime}.

Notably, the Specialized Embedding Models (SEM), which are independently trained for each task, consistently underperform relative to multi-task embedding models. These findings highlight that \mote’s task-specialized expert architecture provides more effective embedding specialization than instruction-conditioning approaches, improving performance across diverse downstream tasks.


\subsection{Inter-task similarity analysis}\label{subsect: dynamic training regime}

Our initial hypothesis was that relying solely on instructions constrains the model's ability to disentangle specialized embeddings, resulting in greater similarity between embeddings across tasks. To test this, we analyze and compare the specialized embeddings produced by \mote and IC when trained on the same data and optimization process. If IC has enough capacity to disentangle the specialized embeddings, increasing the model's specialization capacity (\mote) would not lead to a significant different in the similarities between embedding specializations. On the other hand, if IC constrained the model's capacity to generate disentangled specialized embeddings, training the model with the same data and optimization process but with higher specialization capacity (\mote) will lead to larger differences (lower similarity) between the specialized embeddings.

To test this hypothesis we randomly 128 randomly generic Wikipedia articles\footnote{Dataset available at: \url{https://huggingface.co/datasets/sentence-transformers/embedding-training-data/resolve/main/SimpleWiki.jsonl.gz}} and compute their specialized embeddings for clustering, classification, retrieval-document, and retrieval-query tasks. We then measure the pairwise (inter-task) cosine similarity among these embeddings for both \mote and IC where higher cosine similarity values indicate higher similarity between task-specialized embeddings. To assess statistical significance, we conduct a one-sided Welch’s $t$-test for each pairwise task comparison, following the hypothesis formulated in Expression \ref{eq: statistical hypthesis}, where $\mu^{(T_1, T_2)}_{\mathcal{M}}$ represents the true cosine similarity between tasks $T_1$ and $T_2$ for method $\mathcal{M}$..


\begin{center}
    \begin{align}\label{eq: statistical hypthesis}
    \begin{split}
      H_0: \mu^{(T_1, T_2)}_{\mote}=\mu^{(T_1, T_2)}_{\text{IC}} \\ 
      H_A: \mu^{(T_1, T_2)}_{\mote}<\mu^{(T_1, T_2)}_{\text{IC}}
      \end{split}
    \end{align} 
\end{center} 

\begin{figure*}[ht]
    \centering
        \centering
        \includegraphics[width=\textwidth]{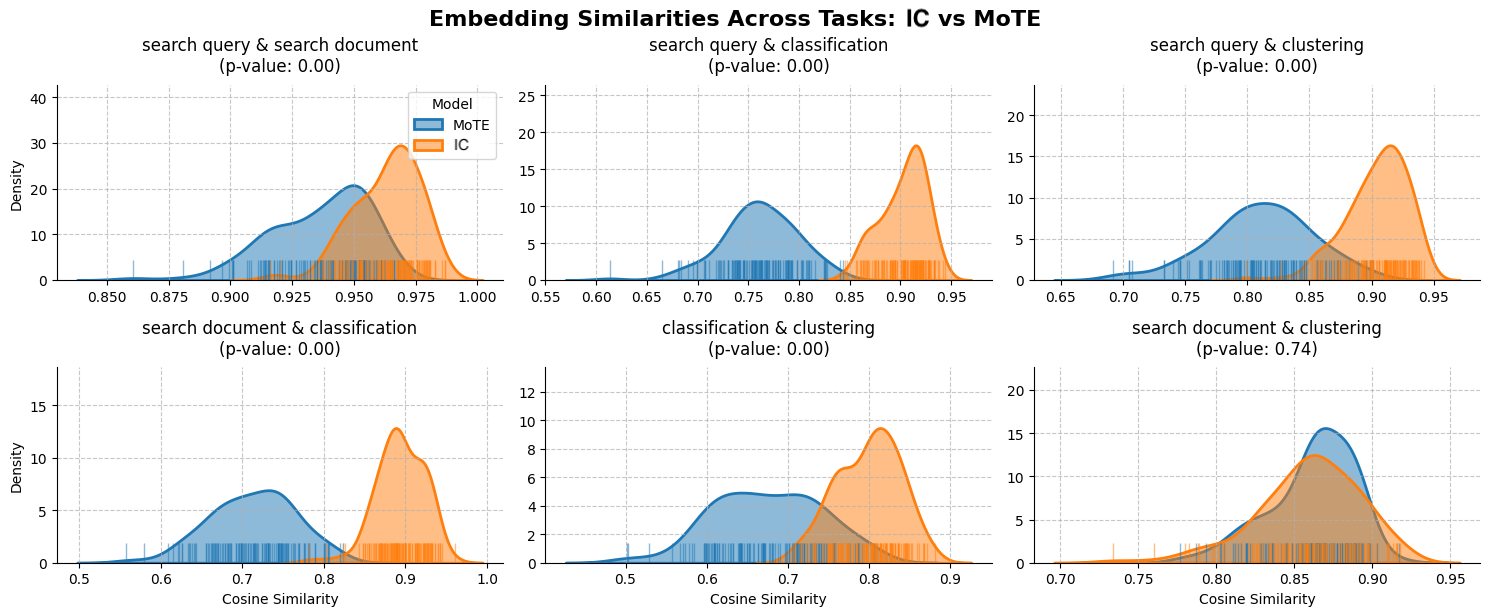} 
        \caption{Inter-task similarity of text representation when using IC and \mote. We observe that in most cases \mote is able to reach a higher degree independence across tasks after the multi-task contrastive learning stage.}
        \label{fig:task-diverging-distributions}
\end{figure*}

Figure \ref{fig:task-diverging-distributions} shows the inter-task similarity distributions for \mote and IC alongside their $t$-test p-value for each of the tasks combinations. We observe that \mote is able to consistently achieves a lower degree of similarity between its specialized embeddings than IC leading to up to $0.3$ lower cosine similarity in some scenarios. These results, alongside the performance improvements in Section \ref{subsection: exp arch perf in scope}, highlights the limitations of IC to tailor embeddings to a specific downstream use and its impact in downstream task performance. Interestingly we observe that on comparison between search document and clustering the inter-task similarity of \mote and IC is not significant different which indicate that the representation for both downstream uses is not significantly different and therefore leveraging the same expert with IC might suffice. We leave the exploration of the consolidation of experts for future research.

\subsection{Generalization to unseen tasks}\label{subsection: exp arch perf out of scope}

Conditioning embedding models on specific instructions to tailor text representations for a finite set of tasks risks creating overly specialized embedding spaces. This risk is further accentuated by \mote, which leverages specialized parameters for each task. This section explores the performance of \mote in tasks that were not used during training, 
namely re-ranking, pair-classification, Semantic Text Similarity (STS), and summarization (we will refer to those as \emph{unseen} tasks). Similar to \cite{nussbaum2024nomic}, we associate unseen tasks to tasks seen during training. Specifically for this experiment we associate re-ranking dataset to the retrieval task and pair classification, STS and summarization datasets to the classification task.

\begin{table}[tb]
  \centering
  \caption{Performance of different methods across tasks not used during training}
  \begin{threeparttable}
  \begin{tabular}{lcccc}
    \hline
    \textbf{Task} & \textbf{EM} & \textbf{SEM} &\textbf{IC} & \textbf{\mote} \\
    \hline
    \hline
    Re-ranking\tnote{1} & \textbf{55.38} & -7.35 & -2.01 & -0.45 \\
    Pair Class.\tnote{2} & 81.52 & -8.22 & -0.8 & \textbf{+0.29} \\
    STS\tnote{3} & 77.91 & -2.82 & +2.11 & \textbf{+2.15} \\
    Summa.\tnote{3} & 32.31 & -5.86 & +1.45 & \textbf{+2.01} \\
    \hline
    Average \tnote{4} & 70.97 & -4.90 & +0.67 & \textbf{+1.25}\\
    \hline
    \hline
  \end{tabular}
  \begin{tablenotes}
\small Comparison of performance gains of instruction-conditioning (IC) and \mote over the non-specialized Embedding Model (EM). \item[1] MAP; \item[2] AP; \item[3] Spearman correlation; \item[4] Score average across 18 datasets.
  \end{tablenotes}
  \end{threeparttable}
  \label{tab:out-of-scope tasks}
\end{table}

Results in Table \ref{tab:out-of-scope tasks} shows that \mote outperforms IC in most cases, achieving the highest gains in STS (+2.15) and Summarization (+2.01), demonstrating its adaptability to unseen tasks. In Pair Classification, \mote maintains a slight advantage (+0.29), while IC shows a small drop (-0.8) with respect to the un-specialized model. Re-ranking is the only task where both approaches underperform relative to EM, likely due to a distribution shift between retrieval training data and the re-ranking evaluation datasets. Notably, several re-ranking datasets include atypical query and passage formats—such as unusually long queries or question deduplication samples—which differ from standard retrieval training scenarios. This suggests that re-ranking benefits from the broader, non-specialized embedding space of EM over task-specialized models. Overall, \mote achieves the highest average performance gains (+1.25), highlighting its effectiveness in generalizing to new tasks while maintaining strong performance.

\subsection{Ablation: Impact of \tacl}\label{subsect: dynamic training regime}

To assess the benefits of \tacl we consider ablate both training regimes when considering the batching strategy and contrastive temperature as task-aware configurations (Appendix \ref{sec:analysis}). 

\begin{table}[tb]
    \centering
    \caption{Comparison of \mote's performance when trained with static or task aware training.}
  \begin{threeparttable}
  \begin{tabular}{lccc}
    \hline
    \textbf{Task} & \textbf{Static} & \textbf{\tacl} \\
    \hline
    \hline
    Retrieval\tnote{1}      & 46.48  & \textbf{47.52}  \\
    Classification\tnote{2}  & 69.00  &  \textbf{69.16} \\
    Clustering\tnote{3}   & 42.82  & \textbf{42.89}  \\
    \hline
    Re-ranking\tnote{4}    & 54.69  & \textbf{54.93}  \\
    Pair Classification \tnote{5}   &  \textbf{81.93} & 81.81  \\
    STS\tnote{6}   & 79.70  & \textbf{80.03}  \\
    Summarization\tnote{6}    & 34.06  & \textbf{34.32}  \\
    \hline
    Avg. dataset performance    & 58.78  & \textbf{59.19}  \\
    \hline
    \hline
  \end{tabular}
  \begin{tablenotes}[flushleft]
\small \tacl employs homogeneous batching for retrieval mini-batches and heterogeneous batching in classification and clustering mini-batches and a contrastive temperature of $0.03$ for retrieval and classification mini-batches and $0.06$ for clustering mini-batches. Static training leverages heterogeneous batching and a $0.03$ contrastive temperature. \item[1] NDCG@10; \item[2] accuracy; \item[3] validity measure; \item[4] MAP; \item[5] AP; \item[6] Spearman correlation.
  \end{tablenotes}
  \end{threeparttable}
  \label{tab:training regimes}
\end{table}

Table \ref{tab:training regimes} shows that \tacl leads to consistent performance improvements of the resulting model with the higher relative gains on tasks in which the regime had to be compromised in the static curriculum such as retrieval where using heterogeneous sampling resulted on $46.48$ but by leveraging homogeneous sampling in \tacl the performance rises to $47.52$. Overall, we observe that \tacl improves average dataset performance by +0.41 across all 56 MTEB datasets. To assess the statistical significance of these gains, we conducted a one-sided Welch’s t-test comparing \tacl and static training outcomes across datasets, obtaining a $p$-value of $1 \times 10^{-4}$, well below the standard $0.05$ threshold. This provides strong evidence that \tacl’s improvements over static contrastive learning are statistically significant.

\section{Ablation: MoTE vs MoE routing}\label{subsect: token level routing}

Unlike MoE's learned Token-Level Routing (TLR), \mote employs Sequence-Level Routing (SLR), directly utilizing task information to select the appropriate expert for processing the entire input sequence. This approach allows \mote to maintain the processing efficiency of IC while training experts on task-specific data. In this ablation study, we compare the effectiveness of SLR against TLR by modifying the routing mechanism in each transformer block while keeping the number of experts constant. To ensure a fair comparison, we augment the input with task-specific instructions (``classification: ", ``clustering: ", ``search query: ", and ``search document: ") so that TLR can also incorporate downstream task information during routing.

\begin{table}[tb]
  \centering
  \caption{Performance of different routing mechanisms}
  \begin{threeparttable}
  \begin{tabular}{p{.62\linewidth}ccc}
    \hline
    \textbf{Task} & \textbf{TLR} & \textbf{SLR} \\
    \hline
    \hline
    Retrieval\tnote{1}       &  45.17 & \textbf{47.52}  \\
    Classification\tnote{2}    & 67.98 &  \textbf{69.16} \\
    Clustering\tnote{3}     &  42.69 & \textbf{42.89}  \\
    \hline
    Reranking\tnote{4}      &  54.40  & \textbf{54.93}  \\
    Pair Classification\tnote{5}   &  80.84 & \textbf{81.81}  \\
    STS\tnote{6}    &  78.47 & \textbf{80.06}  \\
    Summarization\tnote{6}      &  32.67 & \textbf{34.32}  \\
    \hline
    Average\tnote{7}    & 57.86  & \textbf{59.19}  \\
    \hline
    \hline
  \end{tabular}
  \begin{tablenotes}
  \small Performance comparison between MoE's Task-Level Routing (TLR) and \mote's Sequence-Level Routing (SLR). \item[1] NDCG@10; \item[2] accuracy; \item[3] validity measure; \item[4] MAP; \item[5] AP; \item[6] Spearman correlation; \item[7] Average performance across 56 datasets.
  \end{tablenotes}
  \end{threeparttable}
  \label{tab:routings}
\end{table}

Table \ref{tab:routings} shows that SLR consistently outperforms TLR across all tasks, with the largest improvements in Retrieval (+2.35), Semantic Textual Similarity (+1.59), Summarization (+1.65), and Pair Classification (+0.97). By leveraging task information at the sequence level, SLR enables more effective expert specialization, resulting in an overall average performance increase of +1.33 across all 56 MTEB datasets. To assess the robustness of this improvement, we conducted a one-sided Welch’s t-test comparing SLR and TLR results across datasets, obtaining a $p$-value of $3 \times 10^{-2}$, confirming that SLR’s performance gains over TLR are statistically significant at the $0.05$ level.

\section{Ablation: MoE architecture design}\label{subsect: exp eb or eob}

Contemporary MoE literature introduces two strategies for integrating MoE blocks into transformer architectures, each offering a different balance between local and global parametrizations. Specifically, MoE blocks can be incorporated into Every transformer Block (EB) \citep{fedus2022switch} or at Every Other Block (EOB) \citep{lepikhin2020gshard}. In this experiment, we evaluate the performance impact of these design choices when extending embedding architectures with \mote blocks.

\begin{table}[tb]
\centering
  \caption{Performance comparison across different integrations of \mote blocks}
  \begin{threeparttable}
  \begin{tabular}{p{.6\linewidth}ccc}
    \hline
    \textbf{Task} & \textbf{EB} & \textbf{EOB} \\
    \hline
    \hline
    Retrieval\tnote{1}   &  \textbf{47.52}  &  47.20  \\
    Classification\tnote{2}  & 69.16  &  \textbf{69.17}  \\
    Clustering\tnote{3}  &  \textbf{42.89}  &  42.81  \\
    \hline
    Re-ranking\tnote{4}   &  \textbf{54.93}  & 54.91   \\
    Pair Classification\tnote{5}  & 81.81  & \textbf{81.83}   \\
    STS\tnote{6}  & \textbf{80.06}  &  79.96  \\
    Summarization \tnote{6}  & \textbf{34.32} &  33.97 \\
    \hline
    Average\tnote{7}  & \textbf{59.19}  & 59.07  \\
    \hline
    \hline
  \end{tabular}
  \begin{tablenotes}
  \small Comparison between integrating \mote blocks in Every transformer Block (EB) or in Every-Other transformer Block (EOB). \item[1] NDCG@10; \item[2] accuracy; \item[3] validity measure; \item[4] MAP; \item[5] AP; \item[6] Spearman correlation; \item[7] Average performance across 56 datasets.
  \end{tablenotes}
  \end{threeparttable}
  \label{tab:moe arch}
\end{table}

Table \ref{tab:moe arch} reveals that integrating \mote blocks at EB versus EOB leads to minimal performance differences across tasks. While EB achieves a slightly higher average dataset performance (59.19 vs. 59.07), the gains are marginal across retrieval, classification, clustering, re-ranking, and summarization tasks. Notably, EOB slightly outperforms EB in pair classification, whereas EB leads in retrieval, STS, and summarization. These results suggest that while marginal, introducing \mote blocks at EB results in overall better performance compared to doing so at EOB.




\section{Conclusion}

In this paper, we identified the limitations of instruction conditioning for embedding specialization and introduced \mote and \tacl as alternative approaches to overcome these constraints. Our results demonstrate that \mote significantly improves overall performance—particularly in critical tasks such as retrieval—without increasing computational costs, inference latency, or the number of active parameters. By leveraging task-aware contrastive learning and task-based routing, \mote enhances embedding specialization while maintaining computational efficiency.

Additionally, our findings suggest that a hybrid approach—combining \mote with instruction conditioning—could be more effective than either method alone. While \mote improves task disentanglement, we observed that in certain cases, preserving synergies between related tasks is equally important for performance. For example, Table \ref{tab:value of foundational model} shows that \mote negatively impacts clustering performance, while Figure \ref{fig:task-diverging-distributions} indicates that separating clustering from search document tasks does not yield significant benefits. These results highlight the need for architectural modifications that balance task specialization with synergy. We leave the study of these relationships amongst downstream tasks and exploration of hybrid approaches integrating single and multi-task experts for future research.

\section*{Limitations}

While our approach does not increase inference latency or active computational costs, scaling \mote to larger architectures or broader task distributions may introduce new challenges in expert selection and routing efficiency. Future work will explore the generalization of \mote to larger embedding models and further our understanding of task disentanglement and synergy.

By systematically studying these trade-offs, we aim to develop embedding models that maximize specialization while maintaining computational efficiency. Exploring adaptive expert selection mechanisms and dynamic task-aware routing could further enhance the versatility and robustness of multi-task embedding models, paving the way for more effective real-world applications.

\bibliography{bibliography}
\appendix
\newpage
\appendix
\onecolumn
\section{Hyperparameter Sensitivity Analysis}\label{sec:analysis}

\subsection{Batching Strategy}\label{exp: perf vs batching strategy}

The selection of in-batch negative samples is inherently task-dependent, with varying embedding characteristics influencing optimal sampling approaches. Tasks emphasizing \emph{local} semantic nuances, such as Semantic Textual Similarity (STS) \citep{agirre-etal-2012-semeval}, require negatives that capture proximal semantic distinctions. Consider paraphrase identification, where sentences like ``Vivendi shares closed 1.9 percent at 15.80 euros in Paris after falling 3.6 percent on Monday'' and ``in new york, vivendi shares were 1.4 percent down at \$18.29'' demand fine-grained contextual differentiation. Conversely, tasks focused on \emph{global} semantic representations, including classification and clustering, benefit from more diverse negative sampling strategies that capture broader semantic variations.

To empirically validate this hypothesis, we conducted a controlled experiment training an embedding model under two distinct sampling regimes: homogeneous and heterogeneous. By maintaining all other experimental parameters constant, we isolated the impact of sampling strategy.

\begin{figure}[ht]
    \centering
    \begin{minipage}{0.48\textwidth}
        \centering
        \includegraphics[width=0.9\textwidth]{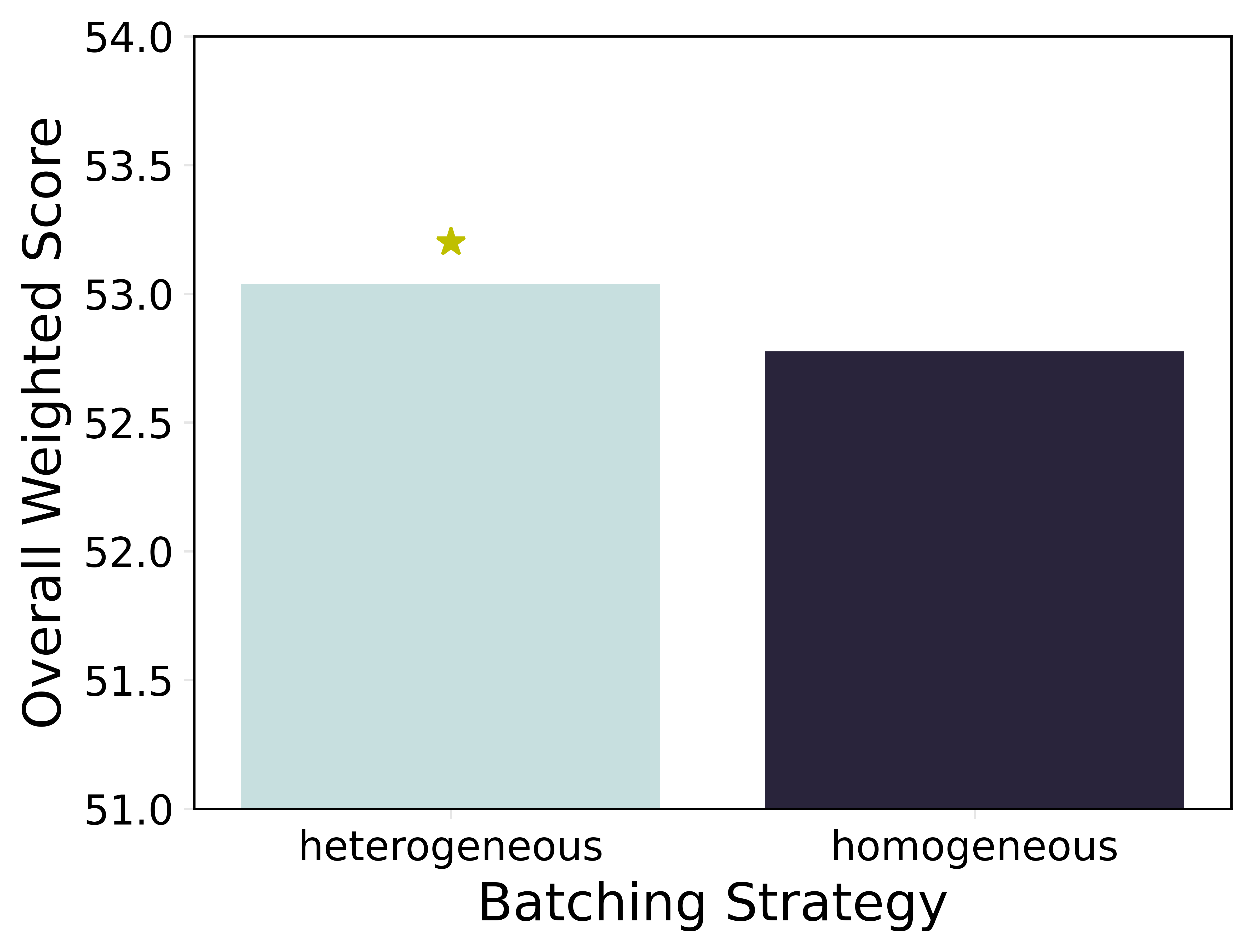}
        \caption{Aggregate Performance: Batching Strategy}
        \label{figure:sampling-overall}
    \end{minipage}\hfill
    \begin{minipage}{0.48\textwidth}
        \centering
        \includegraphics[width=0.75\textwidth]{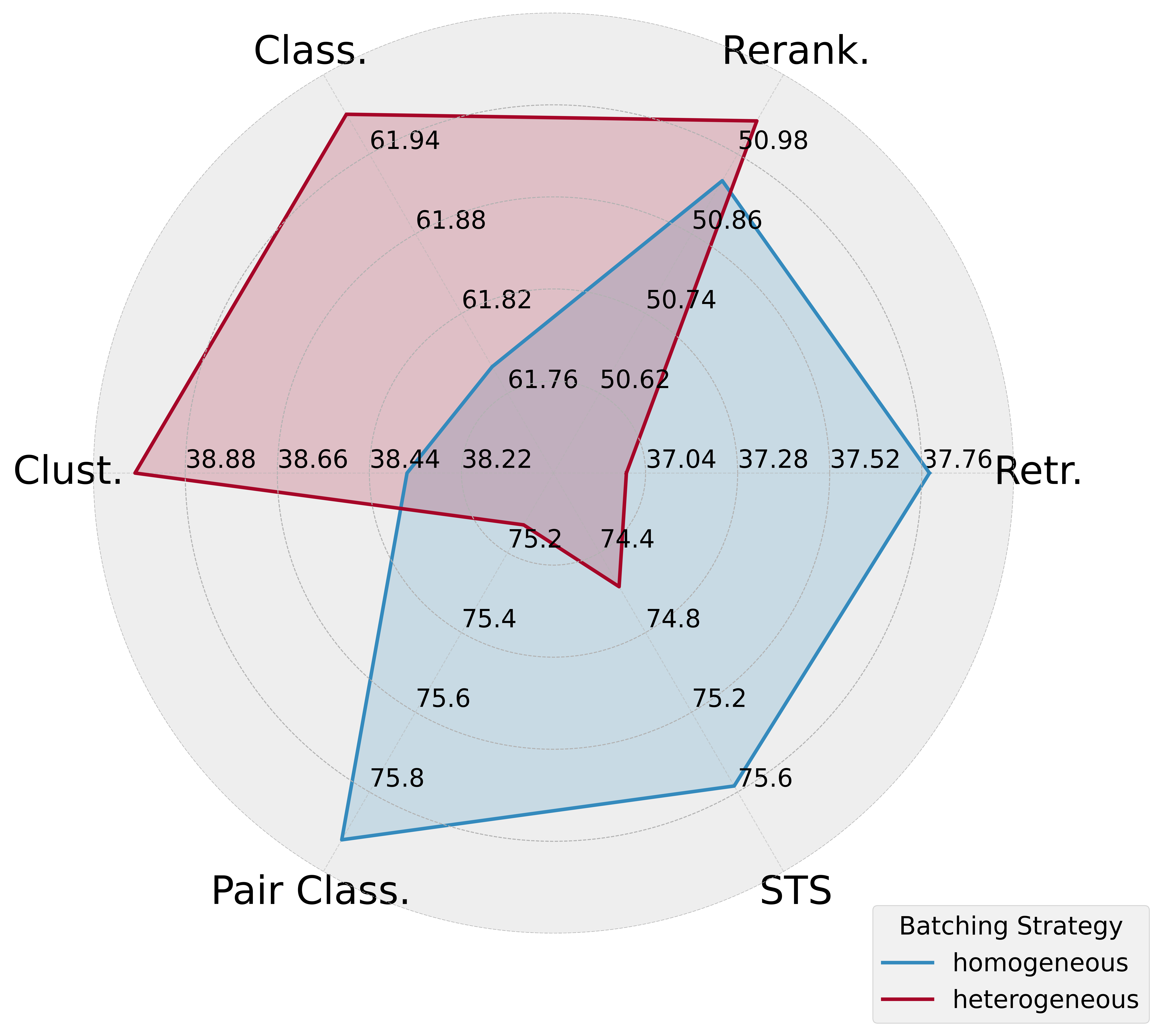}
        \caption{Task-Specific Performance: Batching Strategy}
        \label{figure:sampling-breakdown}
    \end{minipage}
\end{figure}

Figure \ref{figure:sampling-overall} demonstrates that heterogeneous sampling yields the highest aggregate performance metric. However, a granular analysis in Figure \ref{figure:sampling-breakdown} reveals nuanced task-specific variations: homogeneous sampling optimizes performance for local semantic tasks, while heterogeneous sampling proves superior for global embedding objectives. This underscores the critical insight that a universal sampling strategy cannot uniformly optimize performance across diverse downstream tasks.

\subsection{Contrastive Temperature}\label{exp: perf vs cont temperature}

Similar to batching strategies, different contrastive temperatures capture different requirements across downstream task. Lower contrastive temperatures place a higher relative weight on samples that are more semantically similar to the anchor, thus capturing \emph{local} semantic nuances. In contrast, higher temperatures provide a more uniform distribution of the negative weights to capture difference across a more diverse set of negatives. 

To empirically validate this hypothesis, we conducted a controlled experiment training an embedding model under two distinct contrastive temperatures: $0.03$ and $0.06$. By maintaining all other experimental parameters constant, we isolated the impact of contrastive temperature.

\begin{figure}[ht]
    \centering
    \begin{minipage}{0.48\textwidth}
        \centering
        \includegraphics[width=0.9\textwidth]{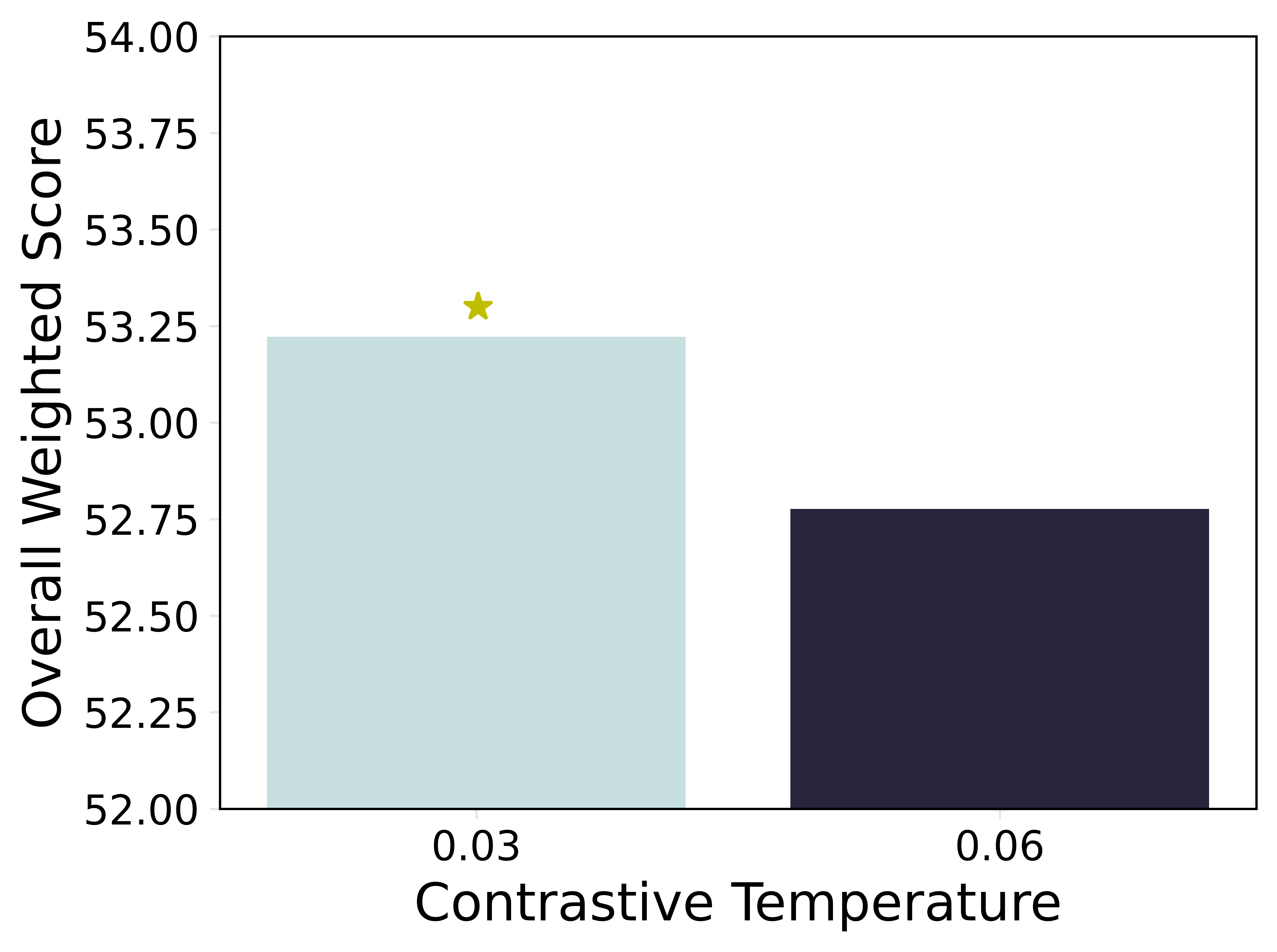} 
        \caption{Overall optimal contrastive temperature.}
        \label{figure: temp-a}
    \end{minipage}\hfill
    \begin{minipage}{0.48\textwidth}
        \centering
        \includegraphics[width=0.78\textwidth]{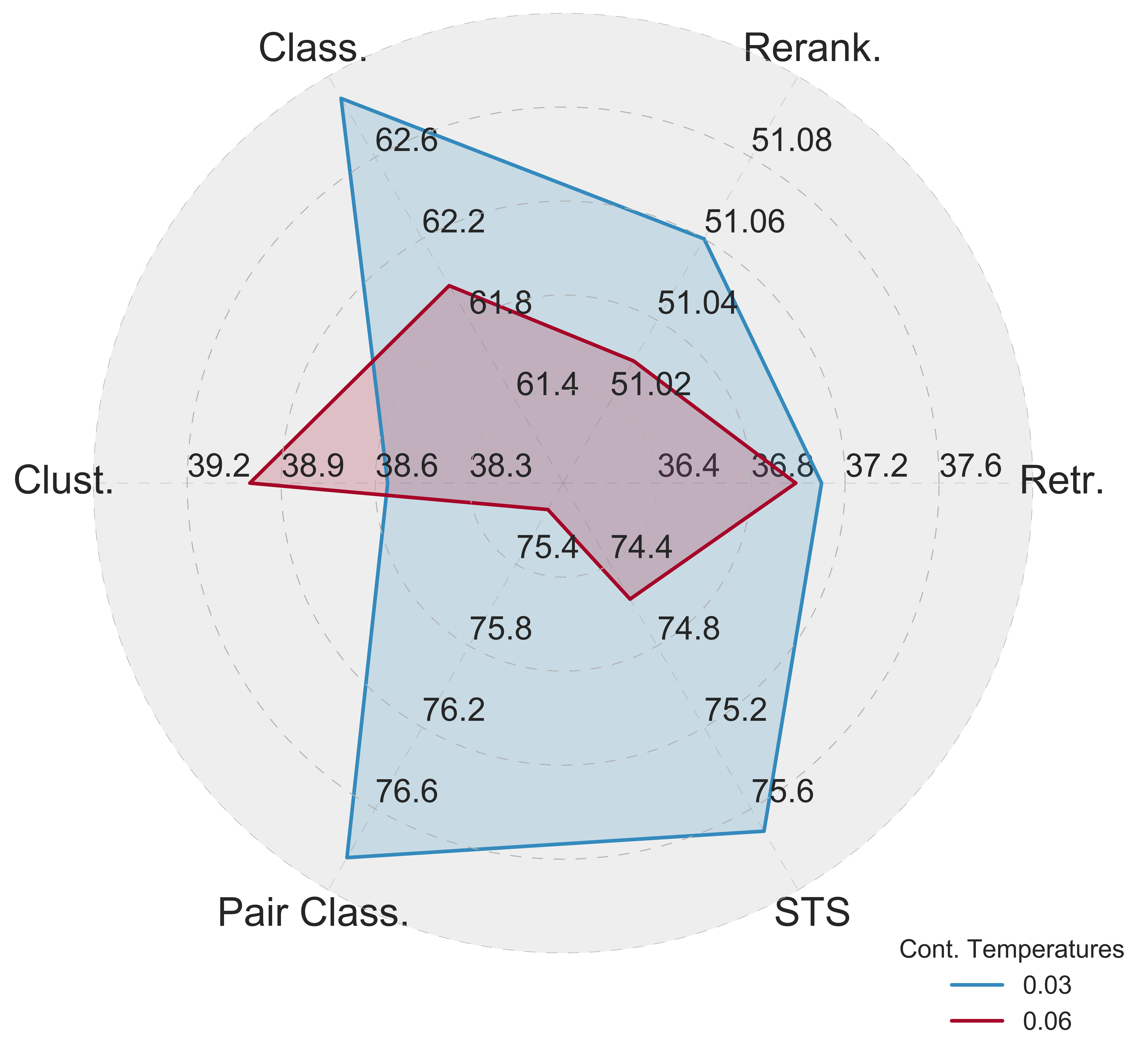} 
        \caption{Optimal contrastive temperature per downstream task.}
        \label{figure: temp-b}
    \end{minipage}
\end{figure}

Figure \ref{figure: temp-a} show that the overall optimal configuration is to choose a temperature of 0.03 but a more detailed analysis in Figure \ref{figure: temp-b} shows that this configuration, while beneficial to the overall performance, hurts clustering tasks.

\section{\ea: Expert Averaging}\label{subsect: tmoe dense converted}

The main drawback of \mote when compared to current dense alternatives is its higher memory consumption which can lead to lower throughput in real world applications such as RAG-indexing or classification which require bulk inference. To alleviate this problem we introduced an Expert Averaging (\ea) compression mechanism to the \mote architecture which alleviated the increased memory footprint at no additional training cost by averaging the expert blocks. In this section we systematically study the performance retention of this technique compared to both \mote and IEM.

\begin{table}[htp]
    \centering
  \caption{\ea performance compared to IEM and \mote across tasks}
  \begin{threeparttable}
  \begin{tabular}{lccc}
    \hline
     & \textbf{IEM}  & \textbf{\mote} & \textbf{\ea} \\
    \hline
    \hline
    Retrieval\tnote{1}  & 45.58 & 47.20    & 46.93  \\
    Classification\tnote{2} & 68.74  &  69.17   &  68.82  \\
    Clustering\tnote{3} &  43.09  & 42.81    &  42.70  \\
    \hline
    Reranking\tnote{4}  & 53.37  & 54.91   & 54.91 \\
    Pair Classification\tnote{5} & 80.72 & 81.83    &  81.59  \\
    STS\tnote{6} & 80.02  & 79.96    &  79.13 \\
    Summarization\tnote{6}  & 33.76 & 33.97    &  33.38  \\
    \hline
    Average dataset performance  & 58.43  & 59.07 &  58.60  \\
    \hline
    \hline
  \end{tabular}
  \begin{tablenotes}
  \item[1] NDCG@10; \item[2] accuracy; \item[3] validity measure; \item[4] MAP; \item[5] AP; \item[6] Spearman correlation
  \end{tablenotes}
  \end{threeparttable}
  \label{tab:wave}
\end{table}

\ea mitigates \mote 's increased memory footprint while retaining $0.17$ average dataset performance over IEM. The average performance retention is largely driven by retrieval and re-ranking tasks with of $1.35$ NDCG@10 and $1.54$ MAP performance improvements of \ea over IEM, respectively.

\section{Evaluation Metrics}\label{appendix: metrics}

\subsection{Normalized Discounted Cumulative Gain at 10}
The Normalized Discounted Cumulative Gain (NDCG) \citep{10.1145/1102351.1102363,wang2013theoretical} is a metric used to evaluate the ranking quality of a search engine or recommendation system by comparing the relevance of retrieved documents to the ideal ranking. The NDCG at position \(10\) is defined as:

\[
\text{NDCG@10} = \frac{\text{DCG@10}}{\text{IDCG@10}}
\]

where:
\begin{itemize}
    \item \(\text{DCG@10}\) (Discounted Cumulative Gain) is calculated as:
    \[
    \text{DCG@10} = \sum_{i=1}^{10} \frac{2^{\text{rel}_i} - 1}{\log_2(i + 1)}
    \]
    \item \(\text{IDCG@10}\) (Ideal Discounted Cumulative Gain) is the DCG@10 of the ideal (perfectly ranked) ordering of results.
    \item \(\text{rel}_i\) represents the relevance score of the result at position \(i\).
\end{itemize}

The NDCG@10 value ranges from \(0\) to \(1\), where \(1\) indicates a perfect ranking.

\subsection{Accuracy}

Accuracy is a metric used to evaluate the overall correctness of a classification model. It is defined as the ratio of correctly predicted instances to the total number of instances:

\[
\text{Accuracy} = \frac{TP + TN}{TP + TN + FP + FN}
\]

where $TP$, $TN$, $FP$, and $FN$ denote the True Positives, True Negatives, False Positives and False Negatives, respectively. The accuracy value ranges from \(0\) to \(1\), with \(1\) representing perfect classification.

\subsection{V-Measure}

V-Measure \citep{rosenberg-hirschberg-2007-v} is a clustering evaluation metric that assesses the quality of a clustering solution by measuring its homogeneity and completeness. It is defined as the harmonic mean of these two metrics:

\[
\text{V-Measure}_\beta = (1 + \beta) \cdot \frac{\text{Homogeneity} \cdot \text{Completeness}}{\beta \cdot \text{Homogeneity} + \text{Completeness}}
\]

where:
\begin{itemize}
    \item \(\text{Homogeneity}\) ensures that each cluster contains only members of a single class. It is defined as:
    \[
    \text{Homogeneity} = 1 - \frac{H(C | K)}{H(C)}
    \]
    \item \(\text{Completeness}\) ensures that all members of a given class are assigned to the same cluster. It is defined as:
    \[
    \text{Completeness} = 1 - \frac{H(K | C)}{H(K)}
    \]
    \item  If $\beta$ is greater than $1$ completeness is weighted more strongly in the calculation, if $\beta$ is less than $1$, homogeneity is weighted more strongly.
    \item \(H(C | K)\) is the conditional entropy of the class distribution ($C$) given the clustering distribution ($K$).
    \item \(H(K | C)\) is the conditional entropy of the clustering distribution given the class.
    \item \(H(C)\) and \(H(K)\) are the entropies of the class distribution and clustering distribution, respectively.
\end{itemize}

$\text{V-Measure}_1$ ranges from \(0\) to \(1\), where \(1\) indicates perfect clustering.

\subsection{Average Precision}

Average Precision (AP) evaluates the model’s ability to rank positive pairs (relevant pairs) higher than negative pairs (non-relevant pairs). It is defined using the ranking of positive pairs:
\[
\text{AP} = \frac{1}{P} \sum_{k=1}^{P} \frac{k}{\text{rank}(k)}
\]
where:
\begin{itemize}
    \item \(P\) is the total number of positive pairs.
    \item \(\text{rank}(k)\) is the rank position of the \(k\)-th positive pair in the sorted list of predictions.
\end{itemize}

AP ranges from \(0\) to \(1\), where \(1\) indicates perfect ranking of all positive pairs above negative pairs.

\subsection{Mean Average Precision}

Mean Average Precision (MAP) is a metric used to evaluate the performance of information retrieval systems, ranking models, or classification tasks with multiple relevance levels. It is defined as the mean of the average precision scores for all queries:

\[
\text{MAP} = \frac{1}{Q} \sum_{q=1}^{Q} \text{AP}(q)
\]

where:
\begin{itemize}
    \item \(Q\) is the total number of queries.
    \item \(\text{AP}(q)\) is the Average Precision for query \(q\).
\end{itemize}

MAP ranges from \(0\) to \(1\), where \(1\) indicates perfect precision across all queries.

\subsection{Spearman correlation}

Spearman’s Rank Correlation Coefficient (\(\rho\)) measures the monotonic relationship between two ranked variables. In the context of STS and Summarization tasks, it evaluates how well the predicted similarity scores or summary scores correlate with human-annotated scores.

\[
\rho = 1 - \frac{6 \sum_{i=1}^{N} d_i^2}{N (N^2 - 1)}
\]

where:
\begin{itemize}
    \item \(N\) is the total number of instances (e.g., pairs of sentences or summaries).
    \item \(d_i = \text{rank}(x_i) - \text{rank}(y_i)\) is the difference between the rank of the predicted score \(x_i\) and the rank of the human-annotated score \(y_i\).
    \item \(\sum_{i=1}^{N} d_i^2\) is the sum of the squared rank differences.
\end{itemize}

Specifically, on STS Tasks \(x_i\) represents predicted similarity scores, and \(y_i\) represents human-assigned similarity scores while in summarization tasks \(x_i\) represents predicted summary quality scores, and \(y_i\) represents human-assigned summary ratings.

Spearman’s \(\rho\) ranges from \(-1\) to \(1\), where \(\rho = 1\) indicates a perfect positive correlation, meaning the predicted rankings match the human-assigned rankings exactly. A value of \(\rho = 0\) signifies no correlation, implying that there is no monotonic relationship between the predicted and human rankings. Conversely, \(\rho = -1\) represents a perfect negative correlation, where the predicted rankings are the exact inverse of the human rankings.

\section{Implementation Details}\label{app: further impl details}

\subsection{Model Architecture}
The MoTE architecture replaces standard transformer blocks with MoTE blocks, each containing four experts assigned to specific instruction types: classification, clustering, search query, and search document. Our implementation builds upon the design proposed by \citep{nussbaum2024nomic}, with two key modifications:

\begin{enumerate}
    \item The standard FeedForward (FF) block is replaced with an \verb|nn.ModuleList| containing multiple FF blocks, one per expert.
    \item The routing mechanism is adapted as described in Section \ref{sec: method}, directing tokens to appropriate experts based on task-specific instructions.
\end{enumerate}

\subsection{Training Procedures}
Mini-batches are constructed by sampling tasks, with data drawn from their corresponding datasets. While the total batch size is kept constant, both the batch construction strategy and the contrastive temperature differ by task type:

\begin{itemize}
    \item Batch Construction Strategy
    \begin{itemize}
        \item For classification and clustering tasks, mini-batches are formed by randomly sampling data points across all datasets associated with the task.
        \item For retrieval tasks, each mini-batch is built by sampling data points from a single dataset, randomly selected for each batch.
    \end{itemize}
    \item Contrastive Temperature:
    \begin{itemize}
        \item A temperature of $0.03$ is used for classification and retrieval mini-batches.
        \item A higher temperature of $0.06$ is applied for clustering mini-batches.
    \end{itemize}
\end{itemize}

\subsection{Batch Size and Infrastructure}

Training is performed with a global batch size of $6144$, made possible using DeepSpeed Stage 2 across a 16-node cluster, with each node equipped with $8$ Nvidia A100 GPUs (40GB). DeepSpeed enables efficient gradient synchronization while broadcasting embedding representations across devices for contrastive loss computation. Notably, gradient caching is not used due to the reliance on in-batch negatives for contrastive learning.

\end{document}